\documentclass[11pt,letterpaper]{article}
\usepackage{geometry}
\usepackage{amsmath}
\usepackage{amssymb}

\usepackage{setspace}   
\usepackage{lipsum} 

\usepackage[pagebackref,breaklinks,colorlinks]{hyperref}

\def\cvprPaperID{*****} 
\def\confName{CVPR}
\def\confYear{2022}

\usepackage[capitalize]{cleveref}
\crefname{section}{Sec.}{Secs.}
\Crefname{section}{Section}{Sections}
\Crefname{table}{Table}{Tables}
\crefname{table}{Tab.}{Tabs.}

\usepackage[utf8]{inputenc} 
\usepackage[T1]{fontenc}    
\usepackage{booktabs}       
\usepackage{amsfonts}       
\usepackage{nicefrac}       
\usepackage{microtype}      
\usepackage{xcolor}         

\usepackage{listings} 

\usepackage{graphicx}

\title{NanoBatch DPSGD: Exploring Differentially Private learning on ImageNet with low batch sizes on the IPU}

\title{NanoBatch Privacy: Enabling fast Differentially Private learning on the IPU}

\author{
  Edward H. Lee\thanks{equal contribution}\\
  Stanford University\\
  \texttt{edward.heesung.lee@gmail.com} \\
   \and
   Mario Michael Krell${}^*$  \\
   Graphcore Inc. \\
   \and
   Alexander Tsyplikhin \\
   Graphcore Inc. \\
      \and
   Victoria Rege \\
   Graphcore Inc. \\
        \texttt{victoria@graphcore.ai} \\
   \and
   Errol Colak \\
University of Toronto \\
   \and
   Kristen W. Yeom \\
   Stanford University \\
   \texttt{kyeom@stanford.edu} \\
}

\begin{document}

\maketitle
\begin{abstract}
Differentially private SGD (DPSGD) has recently shown promise in deep learning. However, compared to non-private SGD, the DPSGD algorithm places computational overheads that can undo the benefit of batching in GPUs. Micro-batching is a common method to alleviate this and is fully supported in the TensorFlow Privacy library (TFDP). However, it degrades accuracy. We propose NanoBatch Privacy, a lightweight add-on to TFDP to be used on Graphcore IPUs by leveraging batch size of 1 (without microbatching) and gradient accumulation. This allows us to achieve large total batch sizes with minimal impacts to throughput. Second, we illustrate using Cifar-10 how larger batch sizes are not necessarily optimal from a privacy versus utility perspective. On ImageNet, we achieve more than $15\times$ speedup over TFDP versus $8\times$ A100s and significant speedups even across libraries such as Opacus. We also provide two extensions: 1) DPSGD for pipelined models and 2) per-layer clipping that is $15\times$ faster than the Opacus implementation on $8\times$ A100s. Finally as an application case study, we apply NanoBatch training for use on private Covid-19 chest CT prediction.

\end{abstract}

\section{Introduction}
Differentially private stochastic gradient descent~\cite{Abadi2016}(DPSGD)
is a technique to train neural networks on sensitive, personal data while providing provable guarantees of privacy. 
Since its introduction~\cite{Abadi2016}, 
recent works have been limited to small datasets and networks in part due to computational challenges.
Overcoming these challenges required a large mini-batch~\cite{McMahan2016}.
However, differential privacy and especially larger mini-batches
impact the privacy loss as well as the accuracy~\cite{Bagdasaryan2019}.
Whereas the original implementation of DPSGD was in TensorFlow~\cite{Abadi2016b,Abadi2016,McMahan2018}(TF)
more recent approaches use JAX with some success~\cite{Anil2021,Subramani2020},
and there are several other approaches to tackle the acceleration on the framework level~\cite{Dupuy2021,Papernot2018,VanderVeen2018}.

This paper explores the application of DPSGD to the ImageNet dataset 
with a ResNet-50~\cite{He2016} 
architecture and analyzes the effect of some parameters on the accuracy, speed, and privacy
on two very different hardware architectures. 
In 2020, Graphcore's second Intelligence Processing Unit (IPU, Mk2) has been introduced.
Some key properties of the Mk2 IPU are that is has 1472 processor tiles, 896MiB on-chip SRAM, 47.5TB/s memory bandwidth, 7.8TB/s inter tile communication,
and that it is a MIMD architecture~\cite{IPU}.
This allows for fine-grained operations on the chip without excessive communication 
to host for fetching weights or instructions.
Instead, in most cases, the instructions and intermediate activations reside on-chip.
Thus in numerous applications, where other acceleration hardware is challenged,
the IPU has shown significant performance advantages like
EfficientNet~\cite{Masters2021}, approximate Bayesian computation~\cite{Kulkarni2020},
multi-horizon forecasting~\cite{Zhang2021}, bundle adjustment~\cite{Ortiz2020},
and particle physics~\cite{Maddrell-Mander2021}.

This paper focuses on processing ImageNet~\cite{Russakovsky2014}.
We use this as a proxy for large scale image processing.
However, there are numerous other applications that can benefit from it, like
brain tumor segmentation~\cite{Li2019,Sheller2019}, cancer detection~\cite{Mukherjee2020},
COVID-19 lung scan analysis~\cite{Covidnpj}, 
and many more in clinical context~\cite{Beaulieu-Jones2018,Geyer2017} 
which have grown in the number of training examples as well as in the size of the data. 
For example, in MRI datasets, there are multiple sequences per patient 
and each sequence contains volumetric data. Larger and more complex models are therefore needed. 

After the introduction of DPSGD, it has also seen increased usage in natural language processing~\cite{Anil2021,Dupuy2021,McMahan2017,Subramani2020}.
A common approach is to actually pretrain on a public dataset without privacy
and then finetune on the privacy sensitive data~\cite{Luo2021}.
Thus larger networks can be trained but the challenge of training on big data
while still respecting privacy is not addressed.

An interesting aspect of image processing are the normalization techniques.
A common approach for ResNet-50~\cite{He2016} is to use batch normalization.
However, batch normalization mixes information across different data samples
and thus violates privacy.
Thus, we will stick to group norm~\cite{Wu2020} in this paper.
Whereas batch norm has an optimal batch size between 32 and 64,
group norm enables high accuracy with much lower batch sizes~\cite{Masters2018}.
Recently, an alternative method, proxy norm~\cite{Labatie2021}, 
has been developed that combines
the benefits of batch norm as the best approach to normalize data and
group norm as the best approach that works even on batch-size of one
and provides the capability of speeding up EfficientNet significantly~\cite{Masters2021}.

This paper addresses acceleration on the hardware level and image processing which has so far been underrepresented in the literature.





\section{Algorithm}

In DPSGD, a batch of images is first sampled randomly at step $t$, per-example gradients are  computed and clipped such that $||g_t||_2 \leq C$, and clipped gradients are accumulated over the entire batch and noised with $\mathcal{N}(0,(\sigma C)^2 I)$, where $C$ AND $\sigma$ are hyperparameters that establish the privacy budget $\epsilon$. The hyperparameters are usually chosen to maximize classification accuracy (utility) for a given $\epsilon$. In our experiments, we fix the total number of epochs and measure both the accuracy and $\epsilon$. We keep $\sigma$ and $C$ constant throughout training and across layers for all experiments. We use the moment accountant implemented in TFDP~\cite{McMahan2018} to compute the noise budget. We clip and inject noise for each gradient example independently to encapsulate DPSGD within an already existing framework, but further throughput gains can be obtained with noising after accumulating. After clipping and noising, gradients are then used to update the parameters via SGD without momentum and with a stepped learning rate decay policy.  

The noised and clipped gradient $g_t$ at step $t$ in the DPSGD algorithm to compute the noised and clipped gradient is shown below. Training examples $x_j$ are sampled randomly from the training set $B$ with batch size $|B|$, per-example gradients are computed and clipped with norm $C$, and the resulting sum is noised with noise multiplier $\sigma^2$:  
$g_t  = \frac{1}{|B|} \left( N(0, \sigma^2 C^2) + \sum_{x_j \in B} \mathrm{clip}(\nabla L(\theta, x_j),C)  \right).$  
In our implementation, we choose to add noise on a per-example basis rather than at the end of the clip-and-accumulate process. This is mathematically equivalent to the original algorithm, provided that the noising of each clipped per-example gradient is sampled from the Normal Distribution with variance $\sigma^2 C^2/|B|$. The purpose of this is to easily integrate DPSGD with the gradient accumulation kernel in the Graphcore TF software stack. Nonetheless, future versions would be easily able to enable support for both methods. Second, this also allows us to clip and noise examples independently with respect to other examples, whether by adaptive clipping or noising strategies \cite{adaclip}. Other frameworks  have limited feature support at the current moment. Our resulting noised gradient is denoted as $\tilde{g}_{t,C}$. It is also worth mentioning that with accumulation, special care needs to be at least considered when fetching the data as indicated in \cite{opacus}.

\begin{align*}
  g_t  &= \frac{1}{|B|} \left(  \sum_{x_j \in B} \tilde{g}_{j,C} \right) 
\text{ where }\tilde{g}_{j,C} =  \left( \mathrm{clip}(\nabla L(\theta, x_j),C) +N(0, \sigma^2 C^2/|B|) \right). 
\end{align*} 


On the IPU, all gradients are clipped and noised on a per-example basis (batch size and microbatch size of 1). Higher per-IPU batch size is supported to support larger microbatch sizes. To achieve a large total (effective) batch size, we perform gradient accumulation and use model replication.   

To compare with IPU experiments, we used public ResNet implementations on TensorFlow and PyTorch (Opacus). We replaced batch norm with group norm to address privacy leakage from batch norm and used public releases of TFDP and Opacus. On the IPU side, we built our experiments around Graphcore's provided public examples repository for CNNs and added the respective code from TFDP for clipping and noising to obtain a DPSGD implementation.\footnote{See latest release in github.com/edhlee/DPSGD-IPU}








\section{Experiments}






\subsection{Hardware comparison}

Multiple publications address challenges in getting differential privacy to run fast.
Hence, we compare two fundamentally different hardware architectures in this section: GPUs and IPUs. 
Thus, we focus on this setting for the comparison of different hardware and SGD and DPSGD. 
Furthermore, we use TensorFlow and TensorFlow Privacy (TFDP) for our baseline comparison. All compared hardware choices have an additional performance hit due to memory constraints by DPSGD 
and being able to run SGD with much higher batch sizes.
Given that Mk2 and A100 share the same lithographic node, we use these to compare. All A100 experiments were performed on the Google Cloud Platform instance
\texttt{a2-highgpu-8g} with 96 vCPUs and 680 GB memory. The total batch size is a result of the single device batch size and the number of replicas for GPUs.
For the IPU, we use a local batch size of 1 
and then use gradient accumulation and replicas for respective larger total batch sizes. 
We compare machines with 8 GPUs to 16 IPUs to match the systems TDP Watts and base packaging setup.

\begin{table}
   \caption{Throughput (img/s) comparison between DPSGD vs SGD on A100 with group norm for different
   $\mu$BS and total batch size combinations with ResNet-50 on the ImageNet dataset.}
    \label{t:A100}
 \centering
  \begin{tabular}{r|rrr|r}
    \toprule
        total & \multicolumn{3}{c|}{DPSGD $\mu$BS}  & SGD  \\ 
        BS & 1  &  2 &  4  & \\ 
    \midrule
        1  & 24 & --  & --  & 28 \\ 
        8  & 62 & 96 & 133 & 201 \\ 
        16 & 43 & 75 & 121 & 315 \\
        32 & 23 & 95 & 82  & 392 \\ 
    \bottomrule
  \end{tabular}
\end{table}

 \begin{table*}[bth]
  \caption{GPU versus IPU throughputs on ResNet-50 (GN32) ImageNet with TensorFlow. The total GPU batch size is denoted as the number of GPUs $\times$ the batch size per GPU, and the IPU batch size is denoted as the number of IPUs $\times$ the batch size per IPU $\times$ the gradient accumulation count. The term 16.32 denotes throughput for a setting with computation in FP16 but master weights in FP32.} 
  \label{t:throughput}
  \begin{center}
  \begin{tabular}{l|l|rr|rr}
    \toprule
    \hline
        total GPU &    total IPU  & \multicolumn{2}{c}{8-A100 GPUs} & \multicolumn{2}{c}{16-Mk2 IPUs} \\ 
        batch size & batch size  & DPSGD & SGD & DPSGD & SGD \\ 
    \midrule
        \hline
          $8\times8$ &  $16\times1\times4$ &  423 & 999  & (4580 16.32)  (5150 16.16) & 6050 (16.16)  \\ 
         $8\times16$ & $ 16\times1\times8$ & 313 & 1758 & (4850 16.32) (5650 16.16) & 6750 (16.16) \\ 
         $8\times32$ &  $16\times1\times16$ & 177 & 2510 & (4900 16.32) (5820 16.16) & 7250 (16.16)  \\ 
         $8\times64$  &  $16\times1\times32$ & OOM&  3046 & (4980 16.32) (5890 16.16) & 7450 (16.16)  \\ 
         $8\times1024$  & $16\times1\times512$ & OOM & OOM & (5650 16.32) (6320 16.16) &  7480 (16.16)\\
    \bottomrule
  \end{tabular}
  \end{center}
\end{table*}

\begin{table*}[bth]
\caption{Throughputs (img/s) of DPSGD on ImageNet across three frameworks. The batch size on Opacus was constrained by memory. Batch size of 8 on TFP gave the highest throughput. Gradient accumulation was not performed on the GPU experiments. ResNet-50 (32 groups) is represented as R50-GN32, and ResNet-101 (32 groups) as R101-GN32. }
\label{tab:table3}
  \begin{center}
\begin{tabular}{|l|l|l|l|r|l|r|}
\hline
Framework   & Hardware & Precision                & Batching      & \multicolumn{1}{l|}{R50-GN32} & Batching & \multicolumn{1}{l|}{R101-GN32} \\ \hline
Opacus & 8xA100 & Mixed Prec. FP16 & 8$\times$64 & 2022 & 8$\times$32 & 830 \\ \hline
Opacus & 8xA100 & FP32                 & 8$\times$64 & 1365 & 8$\times$32 & 665 \\ \hline
Opacus & 8xA100 & TF32                 & 8$\times$64 & 2200 & 8$\times$32 & 845 \\ \hline
TFP    & 8xA100 & Mixed Prec. FP16      & 8$\times$8  & 420  & 8$\times$4 & 205 \\ \hline
NBatch $+$ TFP & 16xMk2   & Float + Stoch. 16.16 & 16$\times$512 & {\color[HTML]{3531FF} \textbf{6320}}  &      16$\times$512    & {\color[HTML]{3531FF} \textbf{3315}}   \\ \hline
\end{tabular}
 \end{center}
\end{table*}

\begin{table*}[bth]
\caption{Throughput of DPSGD training using NanoBatch-TFP on IPUs using ResNet-50 (GN32) on ImageNet. As gradient accumulation is not natively supported by TFP on GPUs, it is not provided below. }
\label{tab:table4}
\begin{center}
\begin{tabular}{|l|l|l|l|l|r|}
\hline
Algorithm & Framework & Hardware & Dtype & Total Batch Size& \multicolumn{1}{l|}{Time per epoch (s)} \\ \hline
DPSGD w/ grad acc & NBatch-TFP     & 16x Mk2 & FP16.16 & 16x4096 &  {\color[HTML]{3531FF} \textbf{205}} \\ \hline
DPSGD w/ grad acc & NBatch-TFP     & 16x Mk2 & FP16.32 & 16x4096 & 228 \\ \hline
DPSGD w/ grad acc & Opacus-PyTorch & 8x A100 & TF32    & 8x8192  & 480 \\ \hline
\end{tabular}
\end{center}
\end{table*}

What is the optimal batch size that maximizes throughput? On the A100 GPU, 
we ran throughput experiments on ResNet-50 (ImageNet) using the TFDP library as shown in Table~\ref{t:A100}. 
For DPSGD, the optimal total batch size that maximizes throughput for the GPU is 8 
independent of the micro-batch size, whereas for SGD, 
throughput is (usually) maximized for larger batch sizes as memory utilization is improved. 
We compare the A100 throughputs to the IPUs. The results are displayed in Table~\ref{t:throughput}.
On GPUs, DPSGD reduces performance by $50$ to $90\%$ depending on the batch size and the use of microbatching.
On IPUs, there is a marginal reduction. 
This is in part because on the IPUs, we use a batch and microbatch size of 1. 
The clip and noising operations add only modest overhead while on the GPU side, 
these operations add significant throughput and memory overheads 
using the TensorFlow vectorized map. 
Table~\ref{tab:table3} compares the throughputs between libraries. 
For the GPUs, the batching configuration was set 
to maximize throughput without grad. accumulation. 
In Table \ref{tab:table4}, accumulation was allowed for both GPUs and IPUs. 
However, TFDP with GPUs does not natively support gradient accumulation and is therefore not shown.

While larger accumulation results in better gradient signal quality, 
smaller accumulation leads to more incremental, but noisier weight updates. 
Furthermore, larger batch sizes increases the privacy budget (TFDP~\cite{McMahan2018}) for a fixed number of epochs. 
The motivation for large batch sizes is to ensure large gradient signal-to-noise ratios with the hope of improving utility. 
However, in order to meet the same privacy budget as an equivalent experiment with smaller batch sizes, the number of epochs must be reduced. Furthermore, papers~\cite{Masters2018} in non-DPSGD research have shown that it is sometimes more useful to perform many noisier gradient updates than fewer steps with extremely large batch sizes. Therefore, there exists the following tension: with low batch sizes, DP gradients would be too noisy to achieve good model training; with high batch sizes, we are constrained with fewer steps to take. 
It is then not too unreasonable to suggest that for a fixed $\epsilon$ budget, 
there exists an optimal batch size that maximizes the accuracy. 
We see this phenomenon in both the IPU and GPU implementations. In the next subsection, we go over the observed effects of batch sizes on the utility and privacy.

\begin{figure*}
  \includegraphics[width=\linewidth]{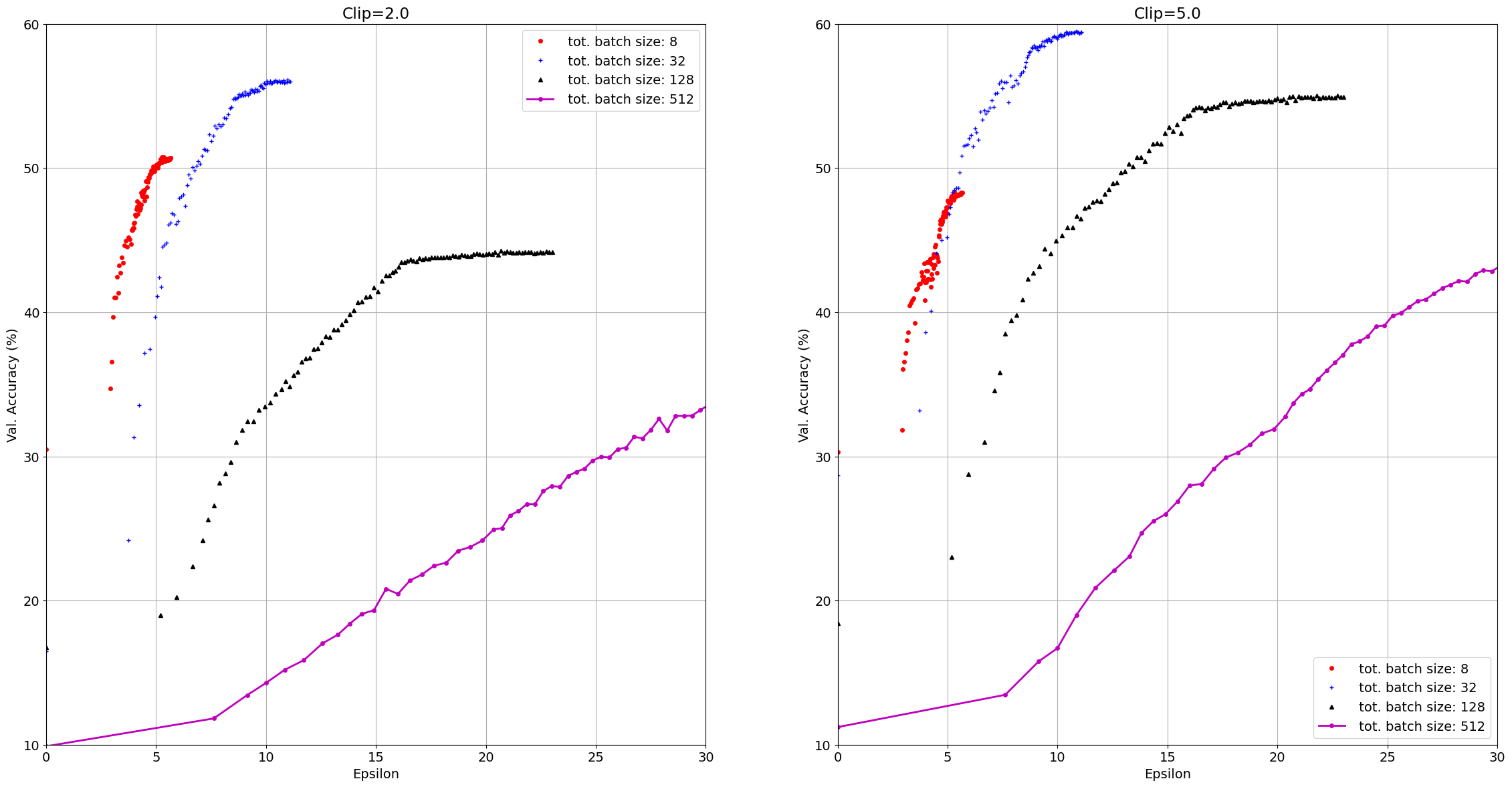} 
  \caption{
 Impact of clipping and batch sizes (with gradient accumulation) on ResNet-18(GN) Cifar-10 accuracy on the IPUs.  With more aggressive clipping, less noise is needed for the same noise multiplier but distorts the gradient signal. With higher batch sizes, the gradient quality improves but increases privacy $\epsilon$. While the quality of the signal improves with larger batches, it does not necessarily lead to better performance as more noisy gradient steps may be better than fewer high quality steps.  
  }
  \label{fig:newfig1}
\end{figure*}

\begin{figure*}
  \centering
  \includegraphics[width=\linewidth]{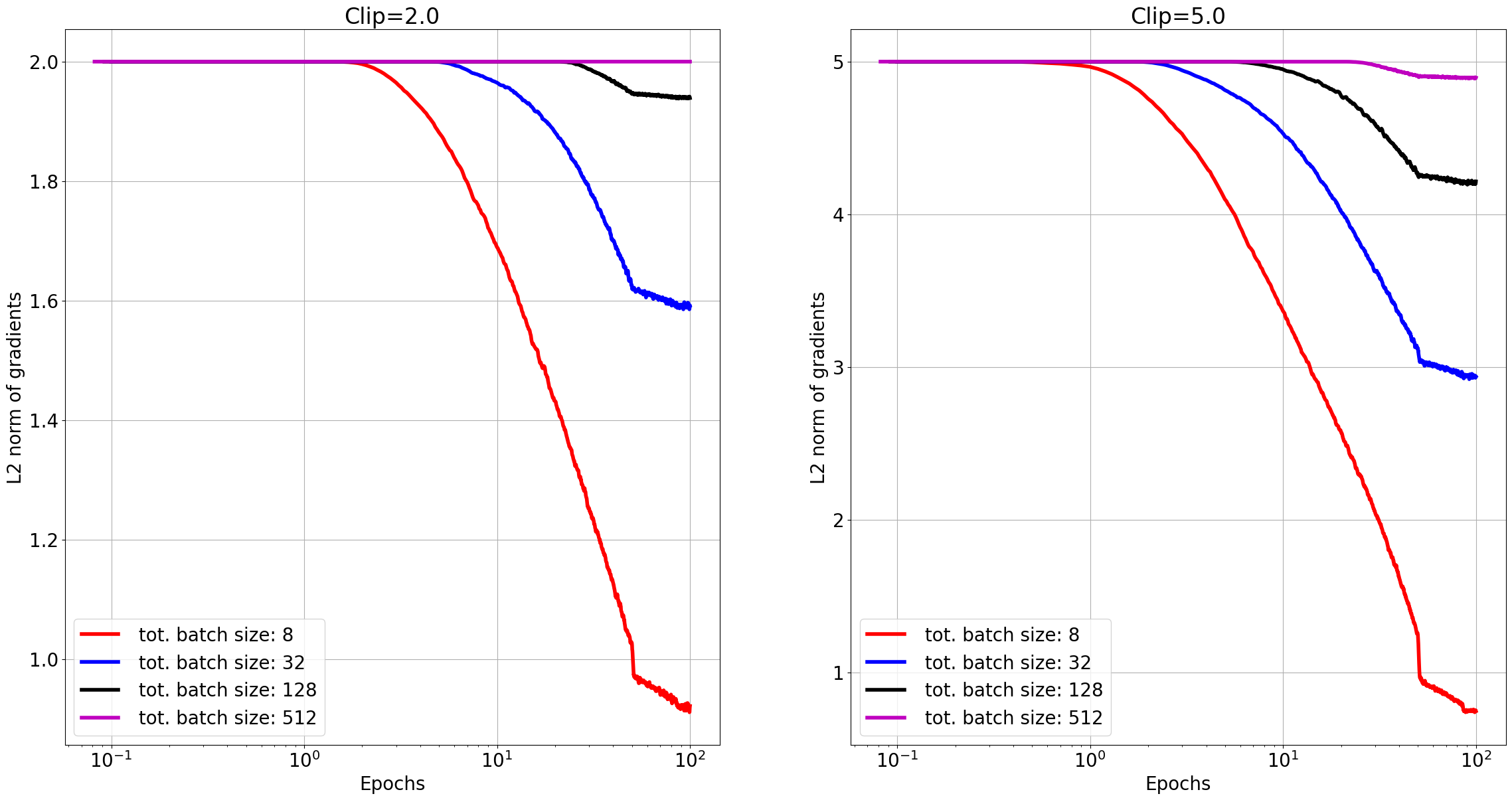} 
  \caption{Gradient norm computed during training for ResNet-18(GN) Cifar-10 with our framework. 
  }
  \label{fig:newfig2}
\end{figure*}

\begin{figure*}
  \centering
  \includegraphics[width=0.5\linewidth]{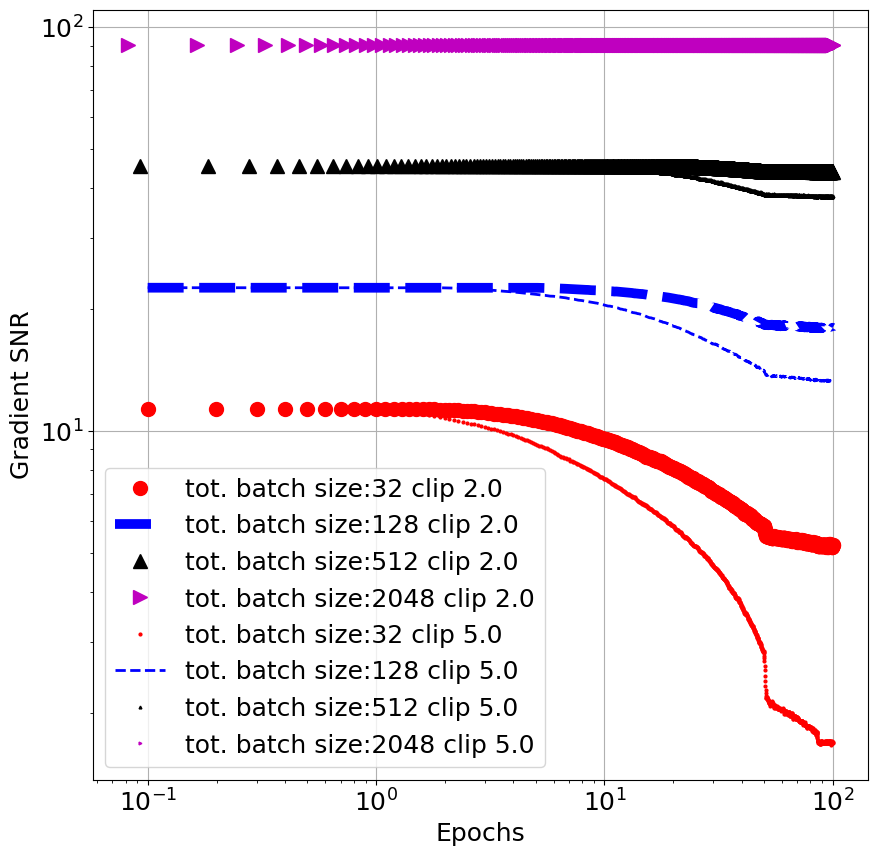} 
  \caption{Gradient SNR computed during training for ResNet-18(GN) Cifar-10 with our framework. Higher SNRs are achieved with larger total batch sizes. As training progresses after epoch 10, the SNR drops significantly. With higher clipping values, the SNR is worse due to the higher injected noise required. 
  }
  \label{fig:newfig3}
\end{figure*}

 \begin{figure*}
  \centering
  \includegraphics[width=0.5\linewidth]{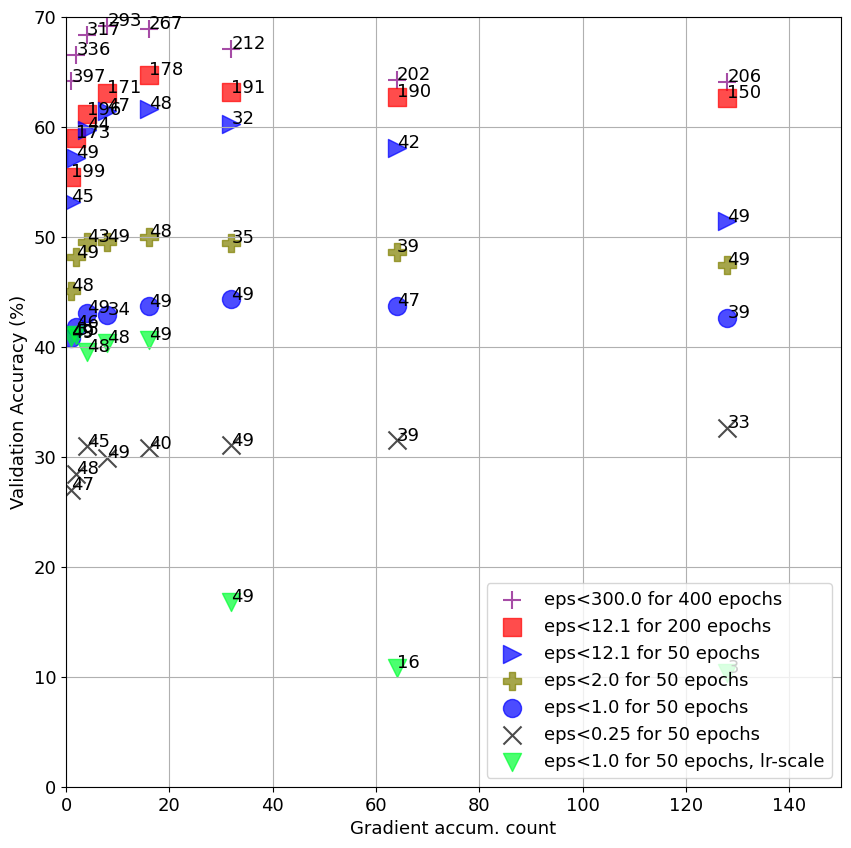} 
  \caption{Plot of the validation accuracy (across epochs) versus gradient accumulation count with an $\epsilon$ constraint using ResNet-18(GN) on Cifar-10 on Opacus with near-identical training parameters to those run on the NBatch-TFDP on IPUs in Fig. \ref{fig:newfig1}. Each of the 7 sets of runs uses a different noise multiplier; the clipping norm is kept the same for all runs. The number by each run is the associated epoch that maximized the accuracy. A larger batch size for a fixed privacy budget is not necessarily optimal in terms of utility. On the other hand, too low of a batch size leads to noisy, and not useful gradient steps. 
  }
  \label{fig:newfig4}
\end{figure*}

\subsection{Setting the batch size for accuracy and privacy}
In the previous section, we show that IPUs running with a batch size of 1 (microbatch size of 1) and with gradient accumulation can accelerate DPSGD with a minor degradation in throughput. Here, we investigate to greater detail the relationship of batch size on utility and privacy on Cifar-10 and ImageNet. 

Using Cifar-10, we illustrate $2\times 4$ sets of training runs (Fig. \ref{fig:newfig1}) with different gradient accumulation counts at two clipping values. In all experiments, we use vanilla SGD with momentum (0.9), the initial learning rate and policies are all the same. From Fig. \ref{fig:newfig1} that plots the accuracy as a function of the privacy loss, it is evident that there exists an optimal batch size; we see that the accuracy peaks at a batch size of around 32 for both clipping values. For these sets of runs, we plot the norm of gradients and their corresponding signal-to-noise ratio (SNR) in Fig. \ref{fig:newfig2} and \ref{fig:newfig3}. From this experiment, we observe the following: with larger batches, the gradient SNR improves, 
provided that the gradient clipping is not too stringent. However, increasing batch size also reduces the number of available gradient steps for a fixed number of epochs and increases the privacy sampling ratio. 

Finally, we replicate the Cifar-10 experiment with NanoBatch using the Opacus Library in Fig.~\ref{fig:newfig4}. 
Here, we show that for a given privacy budget, 
there exists an optimal batch size (gradient accumulation count) that maximizes the validation accuracy. 
The different curves ($\epsilon$) represent different noise multipliers.

\subsection{Learning Rate Scaling}\label{subsubsection:lrscale}

Learning rate scaling \cite{lrscale} is often used to complement with batch size increases. 
By increasing the batch size and thereby improving the confidence of the gradient step, 
the initial learning rate can be increased proportionally to the batch size to speed-up convergence. 
We perform learning rate scaling in our DPSGD experiments for both Cifar-10 and ImageNet. In all of our experiments, we use vanilla SGD with momentum ($0.9$). 

\begin{figure*}
  \centering
  \includegraphics[width=0.5\linewidth]{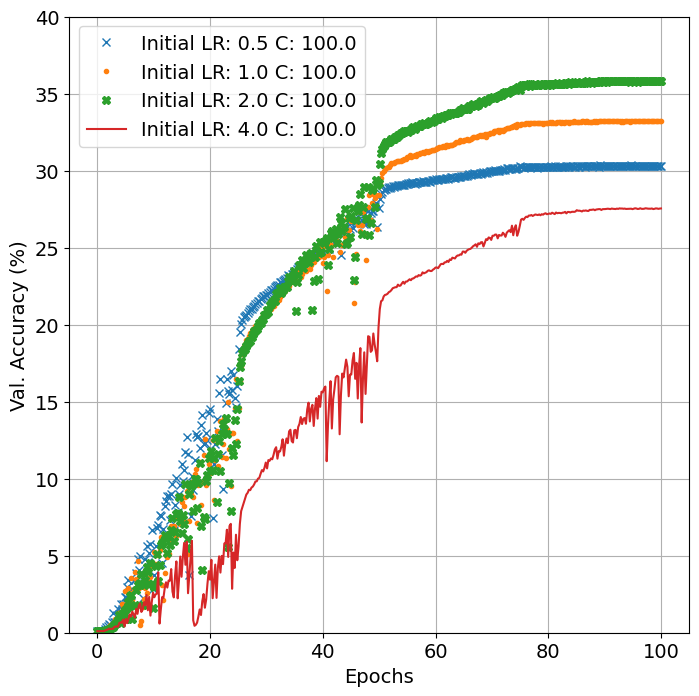} 
  \caption{ImageNet convergence of ResNet-50 with noisy gradients ($z=1.0$, clipping value $C=100$), same total batch size, and different initial learning rates. Convergence is approximately equivalent except for the run with $\texttt{init\_lr}=4.0$.
  }
  \label{fig:newfig4b}
\end{figure*}

On Cifar-10 in Fig. \ref{fig:newfig4}, noise is injected ($\epsilon<1$) where the initial learning rate was set proportionally to the gradient accumulation count ($\texttt{init\_lr}=0.01 \times \texttt{grad\_acc}$). 

without scaling. Compared to the set of runs without scaling the learning rate, the one with scaling had worse utility as gradient accumulation counts $\geq 32$. This suggest that learning rate scaling needs to be used carefully in conjunction with the gradient SNR. We find similar results in ImageNet studies. For example, in Fig. \ref{fig:newfig4b} where the noise multiplier was set to a large value of 1.0, we find that higher initial learning is helpful for large total batch sizes but only up to a certain extent. Too high of learning rates with noisy gradients can lead to overconfident learning steps even with large batch sizes. Further research into the relationship of learning rate policy and noisy gradient policies is warranted. 

\subsection{ImageNet experiments}

\begin{figure*}
  \centering
  \includegraphics[width=0.5\linewidth]{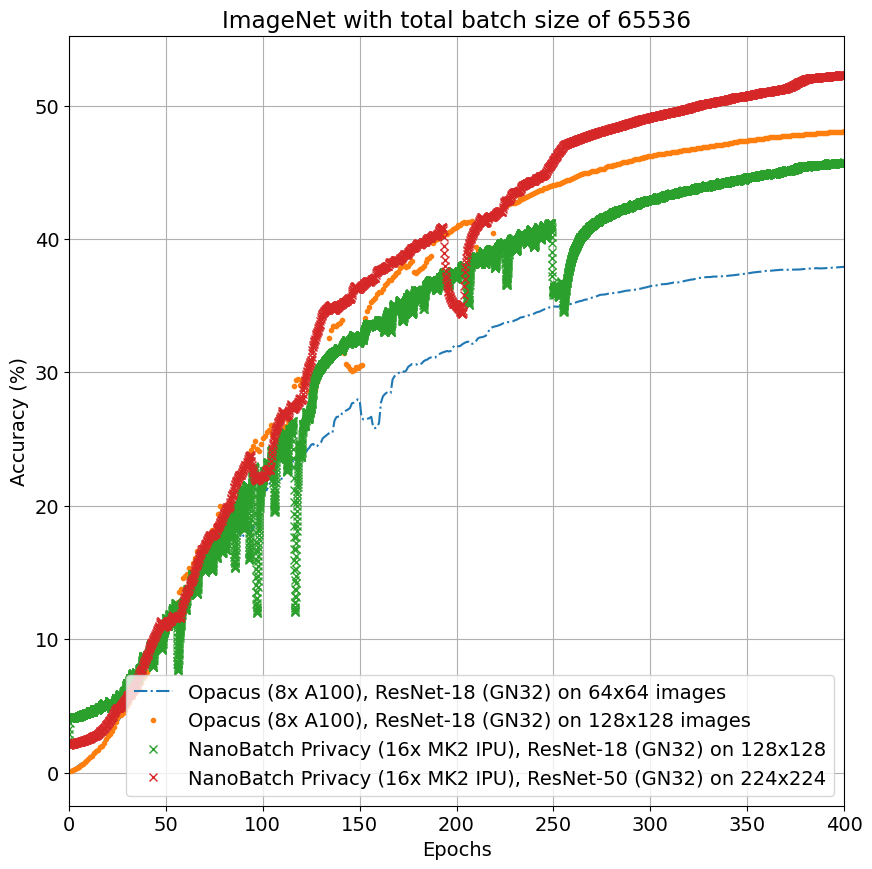} 
  \caption{ImageNet experiments with DPSGD and similar optimization parameters. A noise multiplier of 0.3 was used. Significant gradient accumulation was used. There were out-of-memory issues that arose at seemingly random times during training on the $224\times224$ images with Opacus and results are therefore not shown.   
  }
  \label{fig:newfig5}
\end{figure*}

We demonstrate NanoBatch DPSGD training on ImageNet in Fig. \ref{fig:newfig5}. Informed by our experiments in Section \ref{subsubsection:lrscale}, we choose to set the $\texttt{init\_lr}=1.0$ for all experiments. 
Experiments on GPU with TFP are not shown here due to the extraordinarily long training times. 
Experiments on Opacus with 8 A100s are shown using the same model and optimization parameters consistent with the IPU experiments, 
but with a small image size which was set to 128 and 64 on GPU to speed up experiments. 
With ImageNet training, we set a multiplier of 0.3 and observed 
that the total batch size (using gradient accumulation, microbatch size = 1) 
of $65536$ was needed to achieve good convergence. 
A batch size higher would burn too many examples too quickly, and a batch size lower would offer poor convergence during the warm-up and initial learning phases. With lower noise multipliers, convergence is easily obtained at significantly smaller total batch sizes. 
The model accuracy is slightly lower for batch sizes $\geq 128\times10^3$, and the privacy loss is also higher. We have found that the optimal batch size, which maximizes accuracy with a fixed privacy budget, is significantly greater than those found in experiments using the Cifar-10 dataset.  

To summarize, there exists a tension between large and small batch sizes. First, reduced accuracy with significantly increased batch sizes is shown in experiments even without differential privacy~\cite{Masters2018}. Too high of batch size leads to fewer gradient steps for convergence. Second, decreasing the batch size too much, while keeping noise and clipping constant
the same, results in better privacy sampling ratio and loss but at the cost of noisier gradient steps. From our ImageNet experiments, we see that if the gradients are too noisy, convergence at the beginning of training completely fails. The minimum batch size depends strongly on the gradient signal-to-noise ratio.

We have seen that the optimal batch size can vary widely between datasets. This also applies between applications. For example in NLP, larger batch sizes are more common~\cite{Anil2021}
and can go up to 2 million.
Note that BERT tends to work good with larger batch sizes
but it is more common to stay below $6000$ for fastest 
convergence~\footnote{
\url{https://github.com/mlcommons/logging/blob/master/mlperf_logging/rcp_checker/training_1.1.0/rcps_bert.json}}. 

\subsection{Pipelined models and per-layer clipping}

We also provide support to enable both pipelining and per-layer clipping. Pipelining is a popular method used to fit and train large models across many devices. For example, a feedforward model will be split into $N$ stages where each stage computation is performed on an independent device (IPU). The output activations of each stage is passed as inputs to the next stage. We motivate pipelining as it has become an important part in building ever more larger and complex models with larger datasets.

In order to ensure little degradation in throughput and to prevent limitations in pipelining schemes, we disallow any stage to communicate gradient signals (including gradient norm) to other stages. This means that to enable DPSGD, we clip the per-stage gradients of stage $j$ across $M$ stages independently while ensuring that the original gradient norm bound is respected. Namely, $||g_t||^2_2  = \sum_{j=1}^{M} ||g_{t,j}||^2_2 \leq  C^2 $. In our experiments, we choose a simple uniform partitioning by imposing that each stage has $||g_{t,j}|| \leq C / \sqrt{M}$. This places a tighter constraint on the gradient norm than for the non-pipelined case. 

In NanoBatch DPSGD, we provide support for both pipelined DPSGD and per-layer clipping. Please see our project link \footnote{See latest release in github.com/edhlee/DPSGD-IPU}. With per-layer clipping, we notice speedups of up to $15\times$ compared to the implementation on Opacus using GPUs.

\subsection{Application on AI for sensitive health data}

In our earlier work \cite{Covidnpj}, we demonstrate the ability for AI models to diagnosis Covid-19 disease from non-Covid pneumonia and normal from Chest CT scans across 13 international sites. To the best of our knowledge, our study investigated one of the largest and most diverse patient population. In this work, we use this dataset as one application for NanoBatch DPSGD. 

\begin{figure}[ht]
  \centering
  \includegraphics[width=0.5\linewidth]{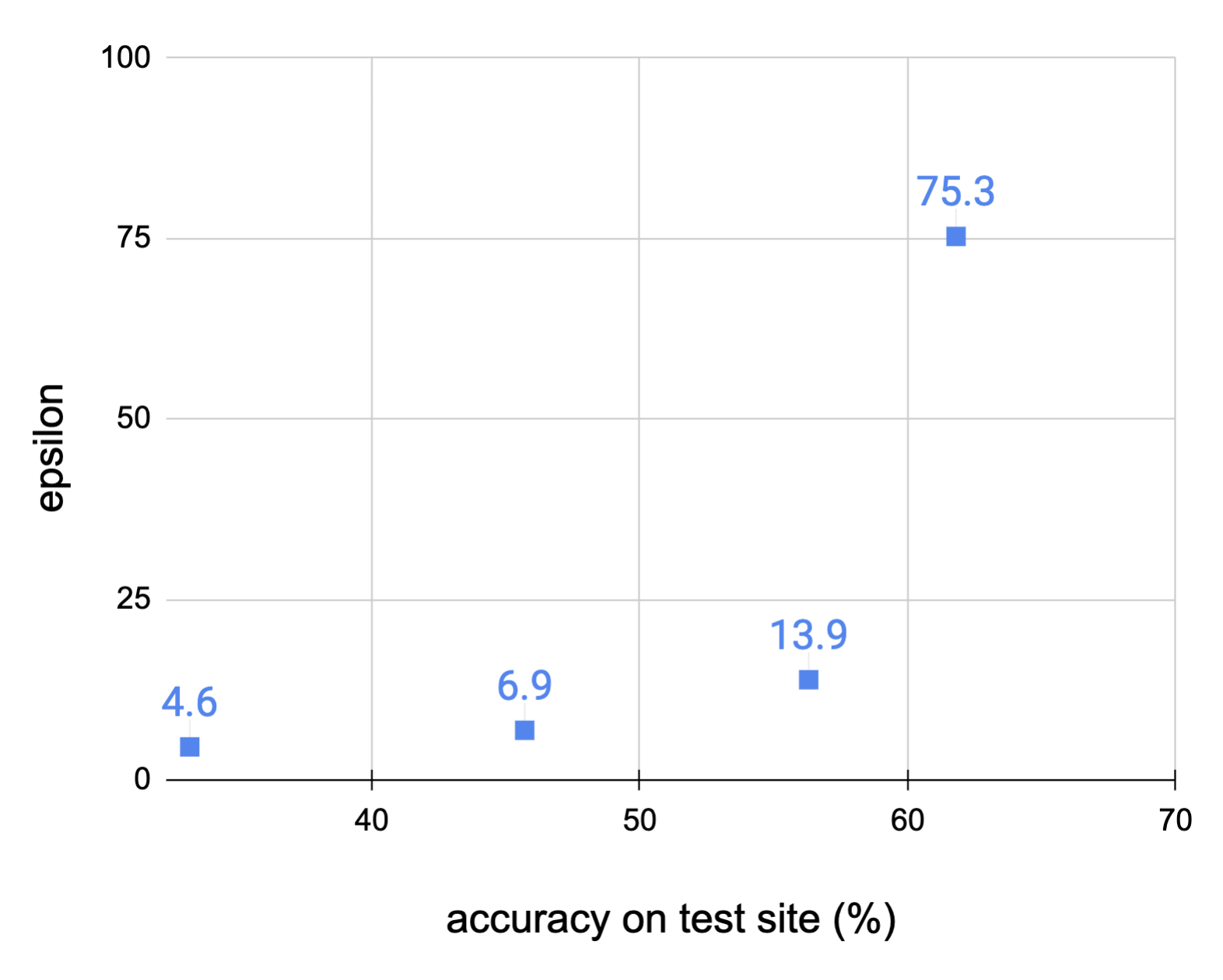} 
  \caption{
   Rapid training using Nanobatch DPSGD applied to Covid-19 diagnosis using Chest CT on 13 international sites. 
  }
  \label{fig:covid}
\end{figure}

In this small study, we apply ResNet-18 (GN32) on 2D images selected from the center of lung across the sagittal axis. There are obvious limitations to 2D such as the scenario when the pneumonia lesions are localized in the sagittal axis. Although the baseline performance of 2D models is lower than that of 3D counterparts, it is smaller and therefore more compact to deploy and share, and faster to train. The training details are posted in the GitHub project. The privacy versus utility for four training runs is shown in Fig.~\ref{fig:covid}. Noise multipliers (0.5 to 2.0) were explored, and the training throughputs ($\approx 4400$ images per second with $8\times$ IPUs) were only marginally off from the baseline throughput in non-DPSGD. Utility can be significantly improved using 3D approaches and data pre-processing.


\section{Conclusion}
In this paper, we propose NanoBatch Privacy, a lightweight add-on to TensorFlow Privacy (TFDP) for Graphcore IPUs. With microbatch and batch size of 1, we are able to achieve near non-private training throughputs. We show using Cifar-10 how privacy loss and utility are impacted by total batch sizes. On ImageNet, we achieve $15 \times$ speedup over GPUs in TensorFlow. Finally, we demonstrate a practical application for fast DPSGD training. Finally, we demonstrate a practical application for fast DPSGD training.


\newpage
\newpage

{\small
\bibliographystyle{ieee_fullname}
\bibliography{references}
}
\end{document}